\documentclass[letterpaper, 10 pt, conference]{support/ieeeconf}
\IEEEoverridecommandlockouts
\usepackage{cite}
\usepackage{amsmath,amssymb,amsfonts}
\usepackage{graphicx}
\usepackage{xcolor}
\usepackage{booktabs}
\usepackage{multirow}
\usepackage[colorlinks,
            linkcolor=black,
            anchorcolor=black,
            urlcolor=black,
            citecolor=black]{hyperref}

\makeatletter
\def\endthebibliography{%
	\def\@noitemerr{\@latex@warning{Empty `thebibliography' environment}}%
	\endlist
}
\makeatother

\begin{document}

\title{\LARGE \bf
Learning Air-Ground Motion Control with Temporal Mode Switching and Cross-Terrain Tracking
}
\author{Ruitian Pang\textsuperscript{1,$\dagger$}, Mingrui Li\textsuperscript{2,$\dagger$}, Xuanting Liu\textsuperscript{1},
        Tiancheng Lai\textsuperscript{1}, Juncheng Chen\textsuperscript{1}, Xiangyu Li\textsuperscript{1},
        Ruibin Zhang\textsuperscript{1},\\ Qishao Wang\textsuperscript{2}, Jin Yu\textsuperscript{4},
        Haiyin Piao\textsuperscript{3}, Fei Gao\textsuperscript{1}, Chao Xu\textsuperscript{1},
        Yanjun Cao\textsuperscript{1,$*$}
        \thanks{This work was supported by National Natural Science Foundation of China under Grant 62103368.}
        \thanks{$^1$The State Key Laboratory of Industrial Control Technology, College of Control Science and Engineering, Zhejiang University, Hangzhou 310027, China.}
        \thanks{$^2$Department of Dynamics and Control, Beihang University, Beijing 100191, China.}
        \thanks{$^3$Jilin University, Changchun, Jilin 130000, China.}
        \thanks{$^4$Harbin Institute of Technology, Harbin 150001, China.}
        \thanks{$^\dagger$Ruitian Pang and Mingrui Li contributed equally to this work.}
        \thanks{Emails: \tt\fontsize{7.8pt}{10pt}\selectfont rt\_pang@zju.edu.cn, atticlmr@gmail.com.}
        \thanks{$^*$Corresponding author: Yanjun Cao (\tt\fontsize{7.8pt}{10pt}\selectfont yanjunhi@zju.edu.cn).}
}

\maketitle

\begin{abstract}
Passive-wheeled terrestrial-aerial bimodal vehicles (TABVs) combine aerial mobility with energy-efficient ground locomotion. However, reliable air-ground mode switching under limited onboard perception and robust ground trajectory tracking across diverse terrains remain challenging when targeting real-world applications. In this work, we propose a learning-based air-ground motion control framework for passive-wheeled TABVs: 1) a learned mode selector for autonomous air-ground motion mode switching. The selector uses historical single-point time-of-flight (ToF) measurements and robot states together with future reference information to determine the active locomotion mode. 2) a reinforcement learning control policy for trajectory tracking. The policy combines proprioceptive observations with future reference information to anticipate trajectory changes. For ground locomotion, multi-terrain training and dynamics randomization enable robust tracking across different terrains. Simulation and real-world experiments demonstrate reliable air-ground switching under limited perception and accurate ground tracking across diverse terrain conditions. The learned selector outperforms a rule-based mode selector in challenging transitions, while the ground controller achieves lower position RMSE than PID across all tested conditions and maintains decent tracking where NMPC fails. With these capabilities integrated, the system tracks a \(101\,\mathrm{m}\) air-ground trajectory through multiple autonomous mode transitions with a position RMSE of \(0.08\,\mathrm{m}\).
\end{abstract}

\section{Introduction}
Terrestrial-aerial bimodal vehicles (TABVs) combine aerial mobility with the energy efficiency of ground locomotion \cite{8741685,9172782,zhang2022autonomous,li2026triphibot}. Passive-wheeled designs use rotor thrust for ground motion without dedicated wheel-driving mechanisms, reducing mechanical complexity \cite{9720983,takahashi2015allround}. Compared with aerial locomotion, ground locomotion generally requires less energy and produces less acoustic noise, making it suitable for long-duration operation. However, reliable air-ground mode switching remains difficult with limited onboard terrain perception \cite{11122320,zhang2022autonomous,fan2019autonomous,wu2023unified}. Variations in wheel-ground interaction across terrains also make it difficult to maintain consistent ground trajectory-tracking performance \cite{atay2020dynamic,10538378}. Therefore, a reliable air-ground control framework is needed for autonomous mode switching and robust ground tracking across diverse terrains.

\begin{figure}[t]
  \centering
  % \vspace{0.2cm}
  \includegraphics[width=\linewidth,keepaspectratio]{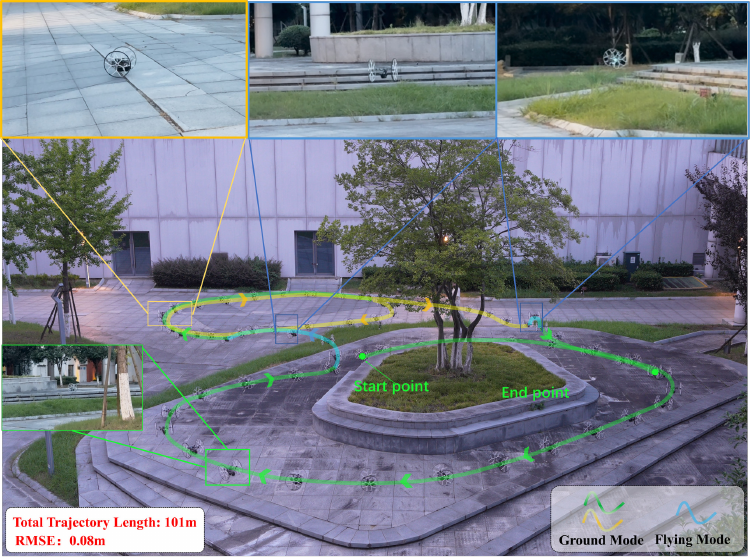}
  \caption{Real-world long-trajectory tracking with autonomous air-ground mode switching. Green and yellow indicate ground locomotion, while blue indicates aerial locomotion. The total trajectory length is $101\,\mathrm{m}$, with a position RMSE of $0.08\,\mathrm{m}$.}
  \label{fig:HeadFigure}
\end{figure}

\begin{figure*}[!t]
\vspace{5pt}
  \centering
  % \vspace{0.2cm}
    \includegraphics[width=0.9\textwidth,keepaspectratio]{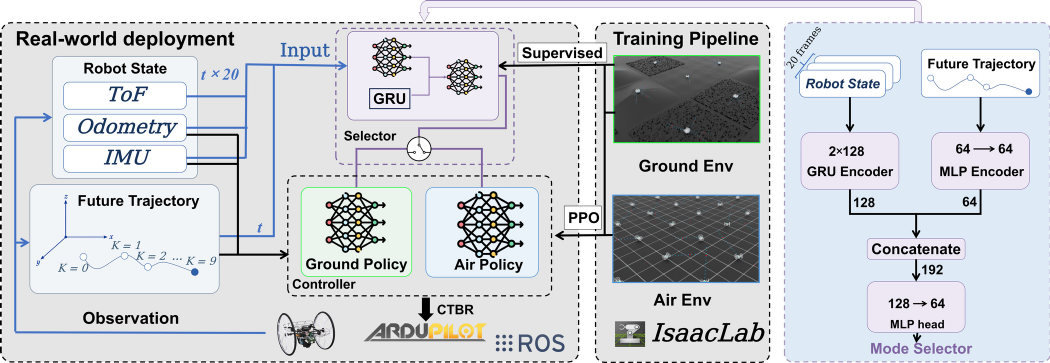}
  \caption{Overview of the proposed framework. The left part shows the real-world deployment pipeline, the middle part illustrates the training environments, and the right part presents the detailed architecture of the mode selector.}
  \label{fig:SystemOverview}
\end{figure*}

Existing mode-switching methods can be divided into rule-based and planning-based approaches. Rule-based methods trigger mode transitions using predefined switching conditions \cite{9341577,zhang2024development}, but depend on manually specified criteria. Planning-based methods determine locomotion modes using environmental perception, mapping, or motion planning \cite{fan2019autonomous,zhang2022autonomous,wu2023unified,11122320}. These methods exploit environmental information but require additional perception and planning computation.

Ground trajectory tracking for passive-wheeled TABVs mainly relies on cascaded, nonlinear, and model-based controllers. Cascaded controllers use cascaded feedback loops for ground trajectory tracking, but rely on simplified ground dynamics and terrain-specific parameter tuning \cite{takahashi2015allround,fan2019autonomous,colmenares2019nonlinear}. Nonlinear and model-based controllers exploit system dynamics to improve tracking performance \cite{colmenares2019nonlinear,atay2021control,zhang2023model,10538378,wang2026tilt}. However, their performance can degrade under model mismatch caused by variations in wheel-ground interaction across different terrains \cite{atay2020dynamic}. In practice, these methods often require terrain-specific modeling or parameter tuning, which limits their adaptability across different terrain conditions.

To address these challenges, we propose an integrated air-ground control framework for passive-wheeled TABVs. The framework integrates autonomous mode switching with aerial and ground trajectory tracking. For mode selection, we design a dual-branch network. The network selects the locomotion mode using historical single-point time-of-flight (ToF) measurements, robot states, and future reference information. It enables online air-ground switching without handcrafted terrain-dependent switching rules or dense terrain representations. For ground locomotion, we train a trajectory-tracking policy using reinforcement learning (RL). The policy leverages robot-state observations and a local reference preview for trajectory tracking. It is trained across diverse terrains with dynamics randomization to improve cross-terrain robustness. The same policy operates without terrain identification or retuning and transfers zero-shot to unseen real-world terrains. The complete system is further validated in a \(101\,\mathrm{m}\) real-world air-ground trajectory with autonomous mode switching, as shown in Fig.~\ref{fig:HeadFigure}.

The main contributions are summarized as follows:
\begin{enumerate}
    \item We develop an air-ground control framework for passive-wheeled TABVs. It enables autonomous switching between independently trained aerial and ground policies through a shared control interface.

    \item We propose a dual-branch mode selector that uses single-point ToF measurements, robot-state history, and future reference information for online air-ground switching without handcrafted terrain-dependent switching rules or dense terrain representations.

    \item We develop a reinforcement learning policy for cross-terrain trajectory tracking using robot-state observations, local reference preview, multi-terrain training, and dynamics randomization, with zero-shot sim-to-real transfer.
\end{enumerate}

\section{Related Work}\label{sec:Related_Work}
\subsection{Mode Switching in Bimodal Robots}
\label{subsec:related_mode_switching}
Existing mode-switching methods can be divided into rule-based and planning-based approaches. Rule-based methods trigger locomotion transitions using predefined conditions. Qin et al. \cite{9341577} use transition commands and height thresholds for mode switching. Zhang et al. \cite{zhang2024development} employ threshold-based switching rules using forward range measurements. These methods are simple to implement, but fixed switching criteria can be sensitive to sensing uncertainty and may require scenario-specific tuning when terrain-induced cues are weak or ambiguous.

Planning-based methods integrate locomotion modes into navigation or trajectory planning frameworks \cite{fan2019autonomous,zhang2022autonomous,wu2023unified,11122320}. Fan et al. \cite{fan2019autonomous} and Zhang et al. \cite{zhang2022autonomous} integrate terrestrial-aerial modes into unified planning frameworks for hybrid mobility. Wu et al. \cite{wu2023unified} jointly optimize trajectories and locomotion modes using NMPC, and Gao et al. \cite{11122320} incorporate terrain perception into bimodal exploration planning. These methods exploit environmental information but require additional perception and planning computation. In contrast, our method determines the active locomotion mode from onboard observations and future reference information without dense terrain representations.
\subsection{Ground Control of Passive-Wheeled TABVs}
\label{subsec:ground_control}
Ground locomotion of passive-wheeled TABVs is affected by wheel-ground interaction and nonholonomic constraints \cite{zhang2023model,atay2020dynamic}. Early works commonly adopt cascaded control structures for ground trajectory tracking. Takahashi et al. \cite{takahashi2015allround}, Fan et al. \cite{fan2019autonomous}, and Colmenares-Vázquez et al. \cite{colmenares2019nonlinear} develop feedback controllers for ground locomotion of aerial-ground robots. Atay et al. \cite{atay2021control} develop a nonlinear controller with sliding-mode control and control allocation. Zhang et al. \cite{zhang2023model} establish a unified nonlinear model for aerial and ground locomotion and employ NMPC for bimodal trajectory tracking. Lin et al. \cite{10538378} also adopt NMPC and validate trajectory tracking on rough and low-friction surfaces. These controllers improve tracking performance through model-based control using simplified representations of wheel-ground contact. However, such simplifications can lead to model mismatch across different terrains and degrade tracking performance.
\subsection{Learning-Based Locomotion Control for Air-Ground Robots}
\label{subsec:learning_locomotion}

Learning-based methods have also been applied to multimodal locomotion control. L'Erario et al. \cite{l2024learning} learn walking and flying behaviors for an aerial humanoid robot and achieve terrain-aware mode switching in simulation. Mandralis et al. \cite{mandralis2025quadrotor} compare reinforcement learning and model-based control for morpho-transition, showing the advantages of learning-based control in agile landing and disturbance recovery. Yin and Pei \cite{yin2026reinforcement} use deep reinforcement learning for air-ground trajectory tracking with a unified policy for flight, ground locomotion, takeoff, and landing. Li et al. \cite{li2026learning} use reinforcement learning to coordinate propellers, active wheels, and tilt actuators over discontinuous terrain. These studies show the potential of learning-based control for multimodal locomotion. However, cross-terrain ground tracking for passive-wheeled TABVs remains less explored, especially without terrain identification or retuning.

\section{Methodology}
\label{sec:methodology}

The proposed TABV control framework is shown in Fig.~\ref{fig:SystemOverview}. We formulate the overall task as reference-trajectory tracking with online locomotion-mode selection and construct task-specific reference representations (Sec.~\ref{subsec:trajectory_preliminaries}). Separate ground and aerial RL policies map robot states and tracking previews to collective thrust and body-rate (CTBR) commands (Sec.~\ref{subsec:motion_control}). A dual-branch temporal selector selects the active policy using recent ToF and robot-state histories together with the current reference intent (Sec.~\ref{subsec:mode_switching}).

\subsection{Problem Formulation and Reference Trajectory}
\label{subsec:trajectory_preliminaries}

We formulate the robot control problem as tracking a time-parameterized reference trajectory with online locomotion-mode selection.

The reference trajectory is represented as an $M$-piece polynomial trajectory, where the $i$-th piece is given by
\begin{equation}
\mathbf{p}^{ref}_i(t)
=
\mathbf{C}_i^{\top}\boldsymbol{\beta}(t),
\qquad
t\in[0,T_i],
\label{eq:reference_trajectory}
\end{equation}
where $\boldsymbol{\beta}(t)=[1,t,\ldots,t^K]^\top$ is the polynomial basis, $\mathbf{C}_i\in\mathbb{R}^{(K+1)\times3}$ is the coefficient matrix, $K$ is the polynomial degree, and $T_i$ is the duration of the $i$-th trajectory segment, with the corresponding reference velocity given by $\mathbf{v}^{ref}_i(t)=\dot{\mathbf{p}}^{ref}_i(t)$. The reference trajectory specifies the desired robot motion and does not explicitly encode the Ground/Air locomotion mode.

At each control step $t$, two task-specific representations are constructed from the reference trajectory: the ground and aerial control policies use a \emph{tracking preview}, which contains future position and velocity information for trajectory tracking, while the mode selector uses a \emph{reference intent}, which captures the vertical evolution of the reference trajectory for locomotion-mode selection.

During offline data collection and annotation for mode-selector training, each rollout sample is assigned a mode label $y_t^*\in\{\mathrm{G},\mathrm{A}\}$ using privileged simulation information and offline labeling rules.

These labels are used during offline data collection and supervised training, but are neither generated nor used during deployment.

\subsection{Reinforcement Learning Control Policies}
\label{subsec:motion_control}

We model ground (\(m=\mathrm{G}\)) and aerial (\(m=\mathrm{A}\)) control as separate MDPs, \(\mathcal{M}_m=(\mathcal{S}_m,\mathcal{A}_m, \mathcal{P}_m,\mathcal{R}_m,\gamma)\), where \(\mathcal{S}_m\), \(\mathcal{A}_m\), \(\mathcal{P}_m\), and \(\mathcal{R}_m\) denote the state space, action space, transition dynamics, and reward function, respectively, and \(\gamma\in[0,1]\) is the discount factor. Each policy \(\pi_{\theta_m}\) maps observations \(o_t^m\) to actions \(a_t^m\) and is optimized to maximize the expected discounted return:
\begin{equation}
\theta_m^{*}
=
\arg\max_{\theta_m}
\mathbb{E}_{\pi_{\theta_m}}
\left[
\sum_{t=0}^{T}\gamma^t r_t^m
\right].
\end{equation}

\subsubsection{Observation Space}

Let \(\mathcal{W}\) and \(\mathcal{B}\) denote the world and body frames, respectively. All position and linear velocity quantities in this subsection are expressed in the world frame \(\mathcal{W}\). Both ground and aerial policies use the observation
\begin{equation}
\mathbf{o}_t^m =
\bigl[
\boldsymbol{\omega}_t^b,
\mathbf{v}_t^w,
\operatorname{vec}(\mathbf{R}_t),
\Delta\mathbf{P}_t^w,
\mathbf{e}_{v,t}^w,
\Delta\mathbf{V}_t^w,
\mathbf{a}_{t-1},
\mathbf{h}_t
\bigr],
\label{eq:observation}
\end{equation}
where \(\boldsymbol{\omega}_t^b\) is the body-frame angular velocity, \(\mathbf{v}_t^w\) is the linear velocity, and \(\mathbf{R}_t\in SO(3)\) denotes the rotation from \(\mathcal{B}\) to \(\mathcal{W}\). The previous normalized policy action is \(\mathbf{a}_{t-1}\), and \(\mathbf{h}_t\) contains the angular velocity, linear velocity, and action history over the previous \(N_h\) steps.

The terms $\Delta\mathbf{P}_t^w$, $\mathbf{e}_{v,t}^w$, and $\Delta\mathbf{V}_t^w$ constitute the tracking preview introduced in Sec.~\ref{subsec:trajectory_preliminaries}. At each control step $t$, the tracking preview is constructed from $N_r$ future samples of the reference trajectory at $t+\tau_i$, $i=0,\ldots,N_r-1$, where $N_r$ denotes the number of preview samples and $\tau_i$ denotes the temporal offset of the $i$-th sample from the current time. The relative reference position is defined as $\Delta\mathbf{p}_{t,i}^w=\mathbf{p}^{ref}(t+\tau_i)-\mathbf{p}_t^w$, where $\mathbf{p}_t^w$ denotes the current robot position in the world frame, and the sampled relative positions are stacked as $\Delta\mathbf{P}_t^w=[\Delta\mathbf{p}_{t,0}^w,\ldots,\Delta\mathbf{p}_{t,N_r-1}^w]$. The velocity information consists of the current velocity-tracking error $\mathbf{e}_{v,t}^w=\mathbf{v}^{ref}(t)-\mathbf{v}_t^w$ and the future reference-velocity changes $\Delta\mathbf{v}_{t,i}^w=\mathbf{v}^{ref}(t+\tau_i)-\mathbf{v}^{ref}(t)$, which are stacked as $\Delta\mathbf{V}_t^w=[\Delta\mathbf{v}_{t,1}^w,\ldots,\Delta\mathbf{v}_{t,N_r-1}^w]$. These quantities provide the policy with both the current tracking deviation and the future motion trend of the reference trajectory.

\subsubsection{Action Space}

The raw policy output is squashed using \(\tanh\) to obtain a normalized action \(\mathbf{a}_t^m\in[-1,1]^4\). The action is converted into CTBR commands by linearly mapping the thrust component to \([T_{\min},T_{\max}]\) and scaling the angular velocity components by a fixed factor \(\omega_{\max}\). Both policies share this CTBR interface for mode switching.

\subsubsection{Reward Function}
\label{sec:rl_control}

The ground reward is defined as
\begin{equation}
\begin{aligned}
r_t^{G}
&=\alpha_1 r_{\mathrm{track}}^{G}
+\alpha_2 r_{\mathrm{speed}}^{G}
+\alpha_3 r_{\mathrm{heading}}^{G}\\
&\quad+\alpha_4 r_{\mathrm{pitch}}^{G}
+\alpha_5 r_{\mathrm{roll}}^{G}
+\alpha_6 r_{\mathrm{smooth}}^{G}.
\end{aligned}
\end{equation}
The tracking terms \(r_{\mathrm{track}}^{G}=\exp(-\|\mathbf{e}_{xy}\|_2/0.5)\) and \(r_{\mathrm{speed}}^{G}=\exp(-(v_{\parallel}-v_{\parallel}^{ref})^2/0.2^2)\) reward horizontal position tracking and speed tracking along the motion direction, respectively, where \(v_{\parallel}\) and \(v_{\parallel}^{ref}\) denote the actual and reference speed components along the motion direction. The heading penalty is \(r_{\mathrm{heading}}^{G}=|\Delta\psi|/\pi\). The attitude penalties \(r_{\mathrm{pitch}}^{G}=\min(\max(|\dot{\theta}|-0.25,0)^2/2^2,1)\) and \(r_{\mathrm{roll}}^{G}=\min(\max(|\phi|-5^\circ,0)^2/(30^\circ)^2,1)\) penalize excessive pitch rates and large roll angles, respectively. The action-smoothness penalty is \(r_{\mathrm{smooth}}^{G}=\min(\|\mathbf{a}_t-\mathbf{a}_{t-1}\|_2^2/0.2,1)\), where \(\Delta\psi\) denotes the heading error, \(\dot{\theta}\) the pitch rate, \(\phi\) the roll angle, and \(\mathbf{a}_t\) the normalized policy action.

The aerial reward is defined as
\begin{equation}
r_t^{A}=\beta_1 r_{\mathrm{track}}^{A}
+\beta_2 r_{\mathrm{heading}}^{A}
+\beta_3 r_{\mathrm{velocity}}^{A}
+\beta_4 r_{\mathrm{smooth}}^{A}.
\end{equation}
The terms \(r_{\mathrm{track}}^{A}=\exp(-\|\mathbf{e}\|_2/0.5)\), \(r_{\mathrm{heading}}^{A}=\exp(-4|\Delta\psi|)\), and \(r_{\mathrm{velocity}}^{A}=\exp(-5|\Delta\psi_v|)\) reward three-dimensional position tracking, heading alignment, and velocity-direction alignment, respectively. Here, \(\mathbf{e}\) denotes the three-dimensional position tracking error, \(\Delta\psi\) the heading error, and \(\Delta\psi_v\) the angular difference between the actual and reference velocity directions. The corresponding action-smoothness penalty is \(r_{\mathrm{smooth}}^{A}=\min(\|\mathbf{a}_t-\mathbf{a}_{t-1}\|_2^2/0.05,3)\). The coefficients \(\alpha_i\) and \(\beta_i\) are positive for reward terms and negative for penalty terms.

\subsubsection{Network Architecture}
The aerial and ground policies use the same asymmetric actor-critic architecture. Both the actor and critic are implemented as three-layer MLPs with hidden dimensions of $[512,256,128]$ and ELU activations.

During training, the actor receives noisy observations to improve robustness to sensing uncertainty, while the critic receives privileged noise-free state information for value estimation.

\subsection{Mode Switching via a Dual-Branch Temporal Network}
\label{subsec:mode_switching}
We formulate air-ground mode selection as a binary classification problem using a state-history branch and a reference-intent branch. The state input at time \(t\) is defined as
\begin{equation}
\mathbf{s}_t =
\left[
d_t,\,
\delta_t,\,
\mathbf{v}_t,\,
\boldsymbol{\omega}_t^b,\,
\mathbf{g}_t^b
\right],
\end{equation}
where \(d_t\) is the normalized single-point ToF measurement. Valid ToF measurements within \(0.03\)--\(20\,\mathrm{m}\) are linearly mapped to \([0,1]\), while invalid measurements are set to zero. The flag \(\delta_t\in\{0,1\}\) indicates validity, with \(1\) for valid returns and \(0\) otherwise. \(\mathbf{v}_t\) is the robot linear velocity expressed in the world frame, \(\boldsymbol{\omega}_t^b\) is the body-frame angular velocity, and \(\mathbf{g}_t^b\) is the projected gravity direction. The state-history branch receives the sequence \(\{\mathbf{s}_{t-k}\}_{k=0}^{N_s-1}\), where \(N_s\) denotes the length of the state-history window.

The reference intent introduced in Sec.~\ref{subsec:trajectory_preliminaries} is defined as
\begin{equation}
\mathbf{i}_t =
\left[
\Delta\mathbf{H}_t,\,
\mathbf{V}_{z,t}^{ref}
\right],
\end{equation}
where $\Delta\mathbf{H}_t=[\Delta z_{t,0},\ldots,\Delta z_{t,9}]$ contains ten relative reference-height changes, with $\Delta z_{t,k}=z^{ref}(t+\tau_k)-z^{ref}(t)$, $k=0,\ldots,9$, where $\tau_k$ denotes the temporal offset of the $k$-th reference sample from the current time $t$, and $z^{ref}(t)$ denotes the vertical component of $\mathbf{p}^{ref}(t)$ in the world frame. The corresponding reference vertical velocities are stacked as $\mathbf{V}_{z,t}^{ref}=[v_z^{ref}(t+\tau_0),\ldots,v_z^{ref}(t+\tau_9)]$, where $v_z^{ref}(t)$ denotes the vertical component of $\mathbf{v}^{ref}(t)$. Together, $\Delta\mathbf{H}_t$ and $\mathbf{V}_{z,t}^{ref}$ constitute the reference intent and characterize the future vertical evolution of the reference trajectory for locomotion-mode selection.

The state branch uses a two-layer GRU to process historical ToF measurements and robot states, while the reference-intent branch uses a two-layer MLP to encode the current reference preview. The resulting features are concatenated and passed to a classification head to produce the Ground and Air logits \(\mathbf{l}_t=[l_{\mathrm G},l_{\mathrm A}]\). A softmax operation converts the logits into mode probabilities.

Only the state branch uses temporal history, while the reference-intent branch uses the current trajectory preview. This prevents outdated reference intent from affecting the current mode decision and keeps the selector responsive to the latest motion command.

\begin{figure}[t]
	\vspace{5pt}
    \centering
    \includegraphics[width=\columnwidth]{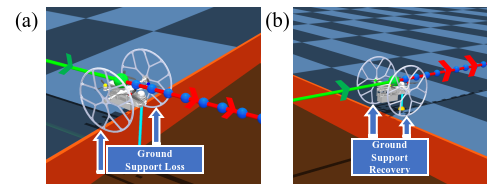}
    \caption{Simulation scenarios for evaluating terrain-induced mode switching: (a) Ground Support Loss and (b) Ground Recovery. The red line denotes the ongoing reference trajectory with blue dots as waypoints.}
    \label{fig:support_scenarios}
\end{figure}

The selector applies asymmetric thresholds with consecutive-frame confirmation to the Air-mode probability \(P_{\mathrm{air}}\), obtained from the output logits via softmax. Ground-to-Air switching requires \(P_{\mathrm{air}}\!\ge\!0.50\) for two consecutive frames, whereas Air-to-Ground switching requires \(P_{\mathrm{air}}\!\le\!0.20\) for eight consecutive frames. These settings are used in all simulation and real-world experiments to suppress mode chattering while favoring rapid Ground-to-Air transitions and conservative returns to Ground.

The mode selector is trained offline by supervised learning using rollout data collected in Isaac Lab\cite{mittal2025isaaclab}. During rollout collection, the target locomotion mode $y_t^*\in\{\mathrm{G},\mathrm{A}\}$ is determined at each step according to predefined scenario-specific labeling rules. For Takeoff, the target mode switches to Air when the reference trajectory is ascending and its height exceeds the local ground level by \(0.02\,\mathrm{m}\). For Ground Support Loss, the target mode switches to Air when the horizontal position of the ToF sensor first reaches the terrain edge. For Landing, the target mode switches to Ground when the reference trajectory is descending, its height above the local ground is below \(0.02\,\mathrm{m}\), and the magnitude of the robot's vertical velocity is below \(0.10\,\mathrm{m/s}\). For Ground Recovery, the target mode switches to Ground when the horizontal position of the robot's body-frame origin first enters the region with restored ground support, without requiring a descending reference trajectory. Representative Ground Support Loss and Ground Recovery scenarios are illustrated in Fig.~\ref{fig:support_scenarios}.

At each rollout step, the frozen ground or aerial expert policy corresponding to the current target mode is executed to track the reference trajectory, and the observations and target mode are recorded. The resulting supervised dataset is represented as \begin{equation} \mathcal{D}=\left\{\left(\mathbf{S}_t,\,\mathbf{i}_t,\,y_t^*\right)\right\}_{t=1}^{N_D},\qquad \mathbf{S}_t=\left\{\mathbf{s}_{t-k}\right\}_{k=0}^{N_s-1}, \label{eq:selector_dataset} \end{equation} where $\mathbf{S}_t$ is the state-history sequence defined above, $\mathbf{i}_t$ is the reference intent, $y_t^*$ is the target locomotion-mode label, and $N_D$ denotes the number of supervised samples. The ToF measurements in $\mathbf{S}_t$ are perturbed during data collection to improve robustness to sensing uncertainty. The target labels $y_t^*$ are generated using the reference trajectory, privileged robot states, and terrain geometry according to the scenario-specific rules above. The selector is trained using a cross-entropy loss to predict the target Ground/Air mode from $\mathbf{S}_t$ and $\mathbf{i}_t$.

\section{Experiments}\label{sec:Experiments}
\subsection{Experimental Setup}

\begin{figure}[t]
	\vspace{5pt}
    \centering
    \includegraphics[width=0.5\columnwidth]{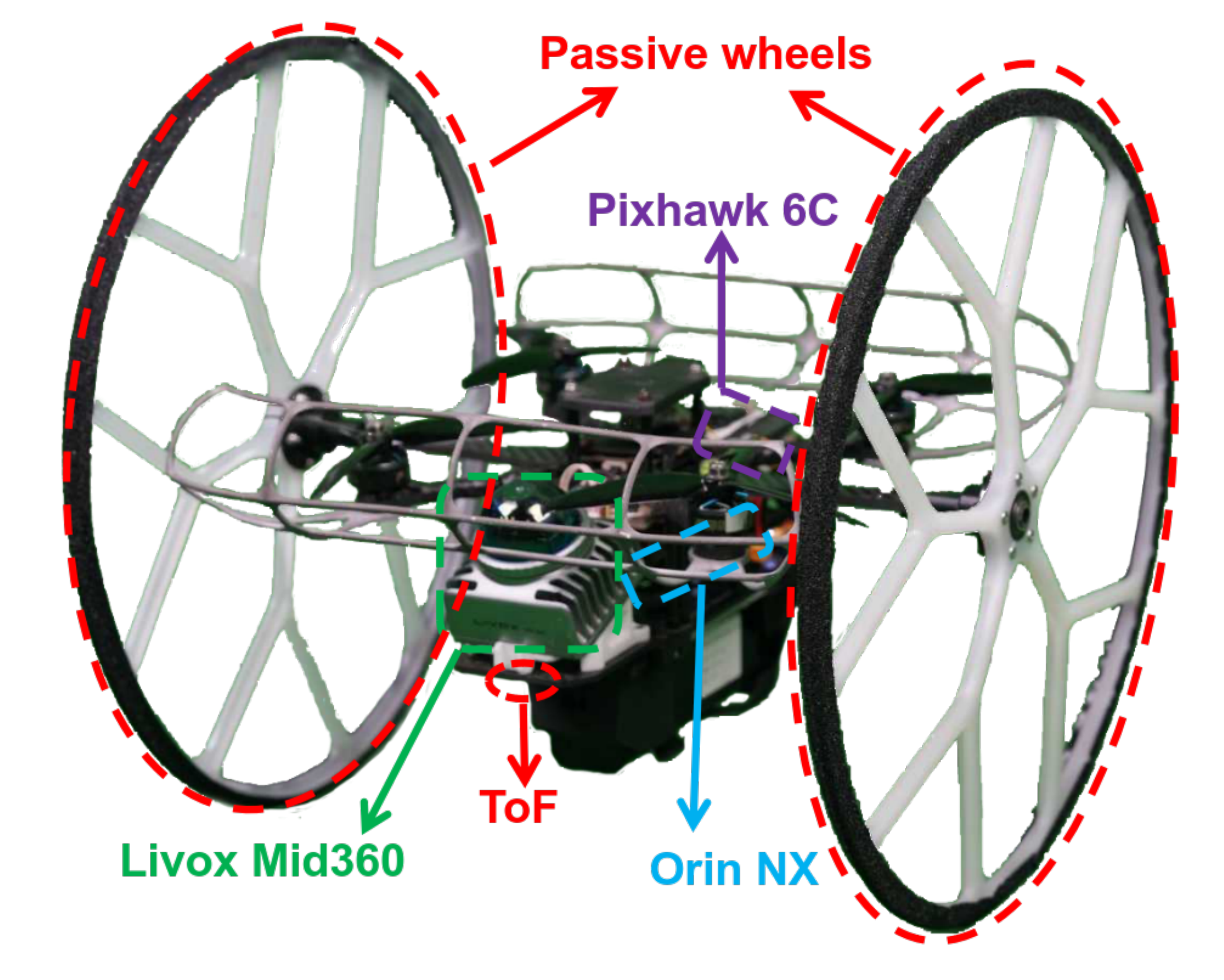}
    \caption{Overview of the customized TABV platform and its onboard sensing and computing modules.}
    \label{fig:tabv_platform}
\end{figure}

Experiments are conducted in simulation and on the customized physical TABV platform shown in Fig.~\ref{fig:tabv_platform}. The aerial and ground control policies are trained in NVIDIA Isaac Lab. The mode selector is trained using expert trajectories collected in Isaac Lab and is evaluated in MuJoCo\cite{todorov2012mujoco} and real-world experiments. The reference trajectories used in the real-world experiments are generated using GCOPTER \cite{9765821}. Simulation experiments are conducted on a workstation equipped with an NVIDIA GeForce RTX 5080 GPU. On the physical platform, localization is provided by FAST-LIO2 \cite{9697912}. The mode selector uses single-point ToF measurements together with onboard state estimates from odometry and IMU. Online inference and control run on an NVIDIA Jetson NX. The ground and aerial control policies are executed at \(200\,\mathrm{Hz}\), while the mode selector runs at \(50\,\mathrm{Hz}\). At each control step, the control policy uses a reference trajectory preview with \(10\) samples. The mode selector uses a \(20\)-frame history window.

\subsection{Evaluation of Learned Locomotion Policies}
\subsubsection{Training Configuration}
\begin{figure}[!t]
	\vspace{5pt}
    \centering
    \includegraphics[width=\columnwidth,keepaspectratio]{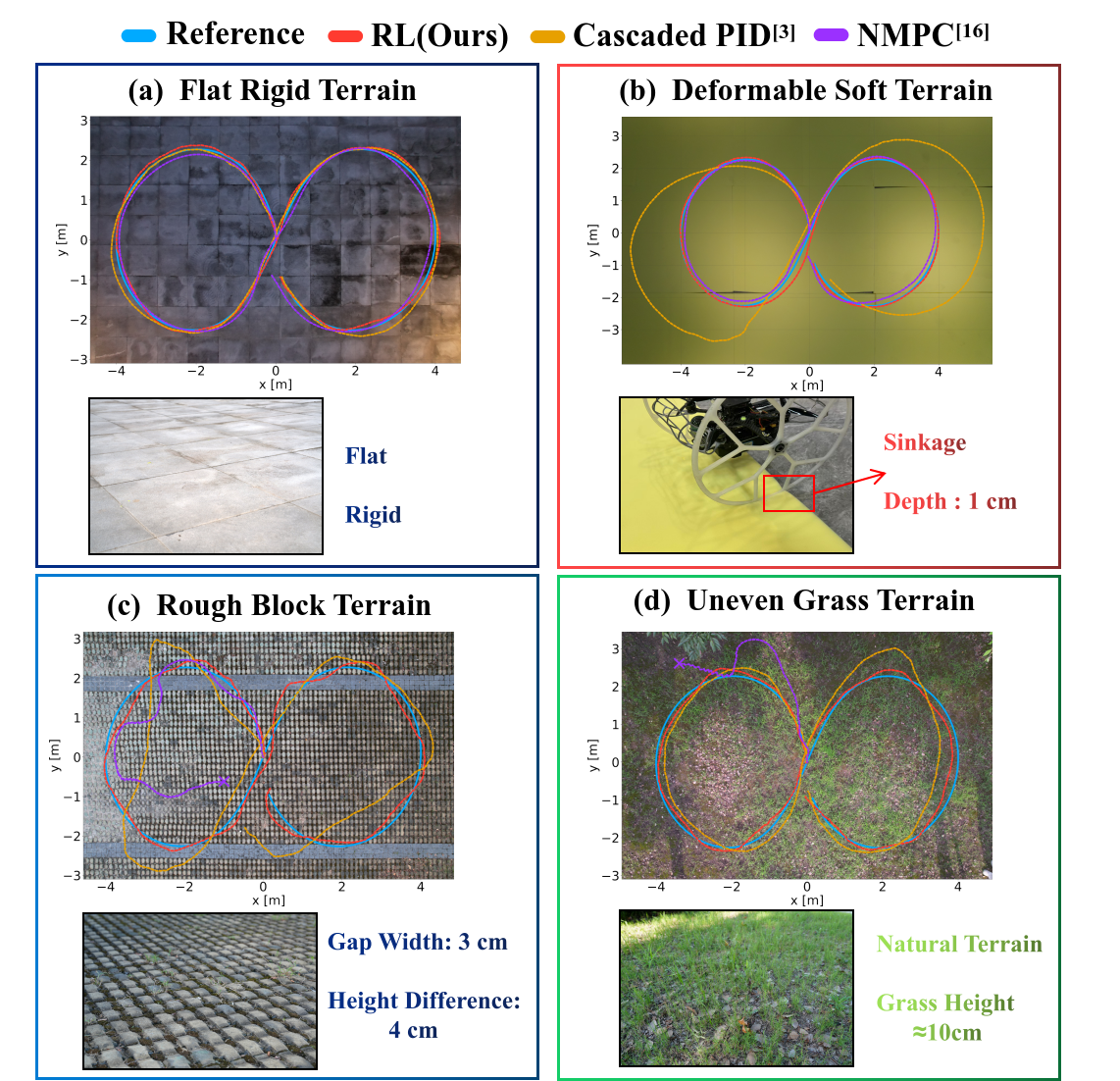}
    \caption{Real-world comparison of ground trajectory tracking on four terrains: (a) flat rigid terrain, $\Delta h\approx0$; (b) deformable soft terrain, with a maximum vertical deformation of $\delta h=1\,\mathrm{cm}$ under the robot weight; (c) rough block terrain, with gap width  $\ell=3\,\mathrm{cm}$ and peak-to-valley height difference $\Delta h=4\,\mathrm{cm}$; (d) uneven grass terrain, with a maximum ground-elevation difference of \(35\,\mathrm{cm}\) across the test area and a grass height of \(h_g\approx10\,\mathrm{cm}\).}
    \label{fig:ground_benchmark}
    \vspace{-0.8em}
\end{figure}

The ground and aerial policies are trained using PPO with 4096 parallel environments for \(1.5\times10^{5}\) and \(2\times10^{5}\) policy steps, respectively. Each policy requires approximately \(40\,\mathrm{min}\) of training.

For ground control policy training, procedurally generated terrains include flat, continuous rough, discrete rough, and terrain-transition regions, with proportions of \(50\%\), \(25\%\), \(20\%\), and \(5\%\), respectively. Continuous rough terrains are generated as random height fields with spatial scales \(\ell\in\{0.1,0.2,0.4,0.8\}\,\mathrm{m}\) and peak-to-valley height variations of \(0.5\)--\(3\,\mathrm{cm}\). Discrete rough terrains use quantized height variations up to \(6\,\mathrm{cm}\), while terrain-transition regions connect flat and rough surfaces.

Reference trajectories are randomly generated during training. Ground trajectories have maximum reference speeds of \(1.9\,\mathrm{m/s}\) and \(1.0\,\mathrm{m/s}\) on flat and rough terrains, respectively, with corresponding maximum curvatures of \(0.75\,\mathrm{m}^{-1}\) and \(0.60\,\mathrm{m}^{-1}\). Aerial trajectories are three-dimensional, with reference speeds of \(0.8\)--\(2.0\,\mathrm{m/s}\), flight heights of \(1.0\)--\(3.0\,\mathrm{m}\), and a maximum curvature of \(0.75\,\mathrm{m}^{-1}\).

To improve sim-to-real transfer, domain randomization is applied during training. The randomized parameters include robot mass, wheel-ground friction coefficients, actuation delays, motor response dynamics, observation noise, and external disturbances.

\subsubsection{Real-World Locomotion Evaluation}

We evaluate the proposed locomotion policies in two real-world experiments: multi-terrain ground trajectory tracking and air-ground hybrid trajectory tracking.

We compare the proposed method with cascaded PID \cite{zhang2022autonomous} and NMPC \cite{zhang2023model} on flat rigid, deformable soft, uneven grass, and rough block terrains using the same figure-eight reference trajectory. Each controller uses fixed parameters across terrains. The proposed method operates without explicit terrain identification.

For quantitative evaluation, we use the position and velocity-vector RMSEs, denoted by \(E_p\) and \(E_v\), respectively: {\small
\begin{equation}
E_p
=
\sqrt{
\frac{1}{N}
\sum_{k=1}^{N}
\left\|
\mathbf{p}_{k}
-
\mathbf{p}_{k}^{r}
\right\|_{2}^{2}
},
\end{equation}
\begin{equation}
E_v
=
\sqrt{
\frac{1}{N}
\sum_{k=1}^{N}
\left\|
\mathbf{v}_{k}
-
\mathbf{v}_{k}^{r}
\right\|_{2}^{2}
},
\end{equation}
} where \(N\) is the number of evaluated samples, and \(\mathbf{p}_k^r\) and \(\mathbf{v}_k^r\) denote the reference position and velocity at time step \(k\), respectively.

As shown in Fig.~\ref{fig:ground_benchmark}, the proposed method tracks the reference at \(1.8\,\mathrm{m/s}\) on flat rigid terrain, \(1.5\,\mathrm{m/s}\) on deformable soft and uneven grass terrains, and \(1.2\,\mathrm{m/s}\) on rough block terrain. Notably, the deformable soft and uneven grass terrains are unseen during training, while the rough block terrain also differs from the procedurally generated training terrains. Moreover, the reference speeds on all three non-flat terrains exceed the maximum \(1.0\,\mathrm{m/s}\) used for rough-terrain training. At these reference speeds, the proposed method reduces \(E_p\) relative to PID by \(92.2\%\), \(77.9\%\), and \(75.8\%\) on deformable soft, uneven grass, and rough block terrains, respectively. Compared with NMPC, it reduces \(E_p\) by \(33.9\%\) and \(17.1\%\) on flat rigid and deformable soft terrains, respectively, while NMPC fails to complete the trajectory on uneven grass and rough block terrains.

Table~\ref{tab:ground_rmse} summarizes the four conditions shown in Fig.~\ref{fig:ground_benchmark} and four additional conditions at lower reference speeds. The proposed method achieves lower \(E_p\) than PID in all eight conditions, with a larger advantage at the higher speed on each terrain. On rough block terrain, for example, increasing the reference speed from \(0.8\) to \(1.2\,\mathrm{m/s}\) raises its \(E_p\) from \(0.131\) to \(0.148\,\mathrm{m}\), compared with \(0.149\) to \(0.612\,\mathrm{m}\) for PID, indicating lower sensitivity to this speed increase. NMPC fails to complete the trajectory on uneven grass and rough block terrains at both tested speeds.
\begin{figure}[!t]
	\vspace{5pt}
    \centering
    \includegraphics[width=0.8\columnwidth,keepaspectratio]{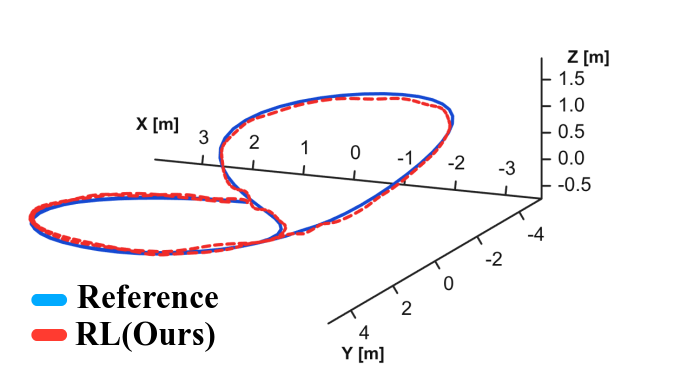}
    \caption{Real-world air-ground hybrid trajectory tracking. The position and velocity RMSEs are \(E_p=0.097\,\mathrm{m}\) and \(E_v=0.252\,\mathrm{m/s}\), respectively.}
    \label{fig:kongdihunhe}
    \vspace{-0.8em}
\end{figure}
\begin{table}[t]
    \centering
    \caption{Real-world ground tracking: \(E_p\) [m] / \(E_v\) [m/s].}
    \label{tab:ground_rmse}
    \renewcommand{\arraystretch}{1.12}
    \setlength{\tabcolsep}{4pt}
    \resizebox{\columnwidth}{!}{%
    \begin{tabular}{lcccc}
        \toprule
        \textbf{Terrain} &
        \begin{tabular}[c]{@{}c@{}}
        \textbf{Max. Ref.}\\
        \textbf{Speed [m/s]}
        \end{tabular} &
        \textbf{Ours} &
        \textbf{PID} &
        \textbf{NMPC} \\
        \midrule

        \multirow{2}{*}{Flat Rigid Terrain}
        & 1.5 & 0.074 / 0.102 & 0.107 / 0.132 & 0.117 / 0.112 \\
        & 1.8 & 0.082 / 0.141 & 0.158 / 0.160 & 0.124 / 0.151 \\
        \midrule

        \multirow{2}{*}{Deformable Soft Terrain}
        & 1.2 & 0.096 / 0.102 & 0.807 / 0.587 &  0.131 / 0.142 \\
        & 1.5 & 0.121 / 0.149  & 1.558 / 0.914 & 0.146 / 0.250  \\
        \midrule

        \multirow{2}{*}{Rough Block Terrain}
        & 0.8 & 0.131 / 0.228 & 0.149 / 0.185 & Failed / Failed \\
        & 1.2 & 0.148 / 0.265 & 0.612 / 0.515 & Failed / Failed \\
        \midrule

        \multirow{2}{*}{Uneven Grass Terrain}
        & 1.2 & 0.132 / 0.180 & 0.508 / 0.230 & Failed / Failed \\
        & 1.5 & 0.178 / 0.149 & 0.807 / 0.914 & Failed / Failed \\

        \bottomrule
    \end{tabular}
    }
\end{table}

We further evaluate the independently trained ground and aerial policies on a hybrid trajectory comprising ground locomotion, takeoff, aerial flight, and landing. As shown in Fig.~\ref{fig:kongdihunhe}, the policies achieve \(E_p=0.097\,\mathrm{m}\) and \(E_v=0.252\,\mathrm{m/s}\) over the complete trajectory. These results demonstrate that the policies can be directly combined through the shared CTBR interface without additional transition-specific control design.

\subsection{Temporal Air-Ground Mode Selector Evaluation}
\subsubsection{Training Configuration}
The selector training data are collected in Isaac Lab using \(4096\) parallel environments, with \(3000\) frames per environment. The network is trained using AdamW for \(30\) epochs with a batch size of \(512\). The initial learning rate and weight decay are set to \(3\times10^{-4}\) and \(1\times10^{-4}\), respectively, with cosine learning-rate decay. The dataset is split into training, validation, and test sets with an \(8:1:1\) ratio at the environment level. To improve robustness to sensing uncertainty, the ToF measurements are augmented with random dropout, Gaussian noise with a standard deviation of \(0.005\,\mathrm{m}\), a sequence-wise bias within \(\pm0.01\,\mathrm{m}\), and a random latency of \(0\)--\(1\) frame. A motion-state dropout probability of \(0.10\) is also applied during training.

\subsubsection{Simulation Switching Performance}

We evaluate the mode selector in MuJoCo under four scenarios: Takeoff and Landing, driven by the reference trajectory, and Ground Support Loss and Ground Recovery, triggered by changes in ground-support conditions. These scenarios cover Ground-to-Air (G$\rightarrow$A) and Air-to-Ground (A$\rightarrow$G) transitions.

The switching delay is defined as
\begin{equation}
T_{\mathrm{switch}}
=
t_{\mathrm{actual}}-t_{\mathrm{expected}},
\label{eq:switch_delay}
\end{equation}
where $t_{\mathrm{expected}}$ is the label transition time determined by the scenario-specific labeling rules in Sec.~III-C, and $t_{\mathrm{actual}}$ is the time at which the corresponding control policy is actually activated. A negative $T_{\mathrm{switch}}$ indicates early switching.

The switching success rate is defined as
\begin{equation}
R_{\mathrm{succ}}
=
\frac{N_{\mathrm{succ}}}{N_{\mathrm{trial}}}\times100\%,
\label{eq:switch_success}
\end{equation}
where \(N_{\mathrm{trial}}\) and \(N_{\mathrm{succ}}\) denote the total number of trials and successful trials, respectively. A trial is successful if the selector completes the required mode transition within the evaluation horizon without an incorrect mode transition. For Ground Support Loss, success additionally requires that the robot never fall more than \(0.05\,\mathrm{m}\) below the pre-event reference height after support loss. Takeoff, Landing, and Ground Recovery are each evaluated over \(50\) independent trials. Table~\ref{tab:selector_switching} reports \(100\%\) success in all three scenarios, with mean switching delays of \(126\,\mathrm{ms}\), \(125\,\mathrm{ms}\), and \(262\,\mathrm{ms}\), respectively.

\begin{table}[t]
	\vspace{5pt}
    \centering
    \caption{Simulation switching performance (delay: mean $\pm$ standard deviation).}
    \label{tab:selector_switching}
    \renewcommand{\arraystretch}{1.1}
    \setlength{\tabcolsep}{5pt}
    \begin{tabular}{lccc}
        \toprule
        \textbf{Scenario} &
        \textbf{Trials} &
        \textbf{Success [\%]} &
        \textbf{Delay [ms]} \\
        \midrule
        Takeoff  & 50 & 100.0 & $126\pm9$ \\
        Landing          & 50 & 100.0 & $125\pm28$ \\
        Ground Recovery  & 50 & 100.0 & $262\pm53$ \\
        \bottomrule
    \end{tabular}
\end{table}

For the Ground Support Loss scenario, we evaluate the selector at five terrain-drop heights, \(h\in\{0.075,0.125,0.175,0.225,0.275\}\,\mathrm{m}\). The TABV follows a \(6.5\,\mathrm{m}\) straight reference trajectory at \(1.0\,\mathrm{m/s}\), with \(50\) trials for each height. Results are summarized in Table~\ref{tab:gsl_height}, with delays reported as mean $\pm$ standard deviation.

\begin{figure}[!t]
	\vspace{5pt}
    \centering
    \includegraphics[width=0.8\columnwidth]{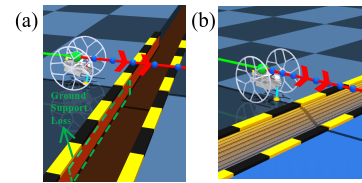}
    \caption{Simulation scenarios for comparison with the rule-based FSM: (a) Gradual Support Loss and (b) Fast Narrow Gap at \(1.8\,\mathrm{m/s}\). The red line denotes the reference trajectory.}
    \label{fig:fsm_scenarios}
\end{figure}

\begin{table}[t]
    \centering
    \caption{Ground Support Loss performance under different terrain-drop heights.}
    \label{tab:gsl_height}
    \renewcommand{\arraystretch}{1.1}
    \setlength{\tabcolsep}{6pt}
    \begin{tabular}{cccc}
        \toprule
        \textbf{Drop $h$ [m]} &
        \textbf{Success/Trials} &
        \textbf{Success [\%]} &
        \textbf{Delay [ms]} \\
        \midrule
        0.075 & 50/50 & 100.0 & $40\pm14$ \\
        0.125 & 50/50 & 100.0 & $25\pm9$ \\
        0.175 & 50/50 & 100.0 & $25\pm9$ \\
        0.225 & 50/50 & 100.0 & $25\pm9$ \\
        0.275 & 50/50 & 100.0 & $26\pm12$ \\
        \bottomrule
    \end{tabular}
\end{table}

Across all \(250\) Ground Support Loss trials, the selector successfully completes the Ground-to-Air transition in every trial, resulting in an overall success rate of \(100\%\) and a mean switching delay of \(28\pm12\,\mathrm{ms}\). The switching delay remains short across all tested terrain-drop heights, indicating that the selector can respond rapidly and reliably to sudden loss of ground support.

\subsubsection{Mode Selector Ablation Study}

To assess the contribution of different selector inputs, we compare \textit{Full} with three ablated variants. \textit{Full} uses 20-frame ToF and motion-state histories together with future reference intent. \textit{w/o ToF} removes the ToF distance and validity inputs, whereas \textit{w/o Intent} removes future reference intent. \textit{Current ToF Only} uses the current ToF measurement while retaining motion-state history and future reference intent. All models are trained from scratch using the same architecture, dataset split, and training configuration. Each model is evaluated over \(100\) trials per scenario in a test batch independent of the main switching evaluation.

As shown in Table~\ref{tab:selector_ablation}, removing ToF reduces the Ground Support Loss success rate from \(100\%\) to \(3\%\) and lowers success rates in Landing and Ground Recovery, highlighting the importance of ToF observations for reliable switching in these scenarios. Removing future reference intent reduces Takeoff success from \(100\%\) to \(0\%\), indicating that takeoff decisions depend strongly on future trajectory information. Using only the current ToF measurement also lowers success rates in Ground Support Loss and Ground Recovery relative to \textit{Full}, showing the benefit of temporal ToF information. Switching delays are averaged over successful trials only and should therefore be interpreted together with the corresponding success rates.
\begin{table}[t]
	\vspace{5pt}
    \centering
    \caption{Ablation results of the mode selector over 100 trials per model per scenario. Switching delays are reported over successful trials only.}
    \label{tab:selector_ablation}
    \renewcommand{\arraystretch}{1.12}
    \setlength{\tabcolsep}{3.0pt}
    \resizebox{\columnwidth}{!}{
    \begin{tabular}{lcccc}
        \toprule
        \textbf{Scenario} &
        \textbf{Full} &
        \textbf{w/o ToF} &
        \textbf{w/o Intent} &
        \textbf{Current ToF Only} \\
        \midrule
        \multicolumn{5}{c}{\textbf{(a) Success Rate [\%]}} \\
        \cmidrule(lr){1-5}
        Takeoff
            & 100.0 & 100.0 & 0.0   & 100.0 \\
        Ground Support Loss
            & 100.0 & 3.0   & 100.0 & 76.0 \\
        Landing
            & 100.0 & 64.0  & 72.0  & 100.0 \\
        Ground Recovery
            & 100.0 & 25.0  & 92.0  & 73.0 \\
        \midrule
        \multicolumn{5}{c}{\textbf{(b) Mean Switching Delay [ms]}} \\
        \cmidrule(lr){1-5}
        Takeoff
            & 103 & 75  & -- & 110 \\
        Ground Support Loss
            & 31  & 265 & 21     & 29 \\
        Landing
            & 107  & 146 & 15     & 86 \\
        Ground Recovery
            & 289 & 101 & 470    & 227 \\
        \bottomrule
    \end{tabular}}
\end{table}
\subsubsection{Comparison with a Rule-based mode selector}
We compare the learned and rule-based selectors in the Gradual Support Loss and Fast Narrow Gap scenarios shown in Fig.~\ref{fig:fsm_scenarios}, with the latter tested at \(1.8\,\mathrm{m/s}\). The former progressively reduces ground support, whereas the latter requires rapid mode switching. Both methods use the same ToF stream, \(50\,\mathrm{Hz}\) decision frequency, two-frame confirmation, simulation initializations, and ToF perturbations. The rule-based selector uses the ToF distance recorded with the robot initially level on the ground as a fixed baseline and switches to Air when the ToF return is invalid or exceeds this baseline by \(0.04\,\mathrm{m}\) for two consecutive frames. A trial fails if the robot falls more than \(0.05\,\mathrm{m}\) below the pre-event reference height. All thresholds are tuned once and then fixed for testing. Each method is evaluated over \(50\) trials per scenario. The learned selector achieves success rates of \(92.0\%\) and \(96.0\%\), compared with \(54.0\%\) and \(56.0\%\) for the rule-based selector, demonstrating more reliable switching in both scenarios.
\begin{table}[t]
    \centering
    \caption{Comparison with the rule-based selector under challenging terrain-induced switching scenarios.}
    \label{tab:fsm_comparison}
    \renewcommand{\arraystretch}{1.1}
    \setlength{\tabcolsep}{6pt}
    \begin{tabular}{lcc}
        \toprule
        \textbf{Scenario} &
        \textbf{Method} &
        \textbf{Success [\%]} \\
        \midrule
        \multirow{2}{*}{Gradual Support Loss}
        & Learned & 92.0 \\
        & Rule-based & 54.0 \\
        \midrule
        \multirow{2}{*}{Fast Narrow Gap}
        & Learned & 96.0 \\
        & Rule-based & 56.0 \\
        \bottomrule
    \end{tabular}
\end{table}
\subsubsection{Real-World Mode Switching Evaluation}
Following simulation evaluation, we directly deploy the trained mode selector on the physical TABV without fine-tuning. In a stair-like environment with large height variations, the robot follows a reference trajectory and autonomously switches between ground and aerial locomotion.

\begin{figure}[t]
	\vspace{5pt}
    \centering
    \includegraphics[width=\columnwidth,keepaspectratio]{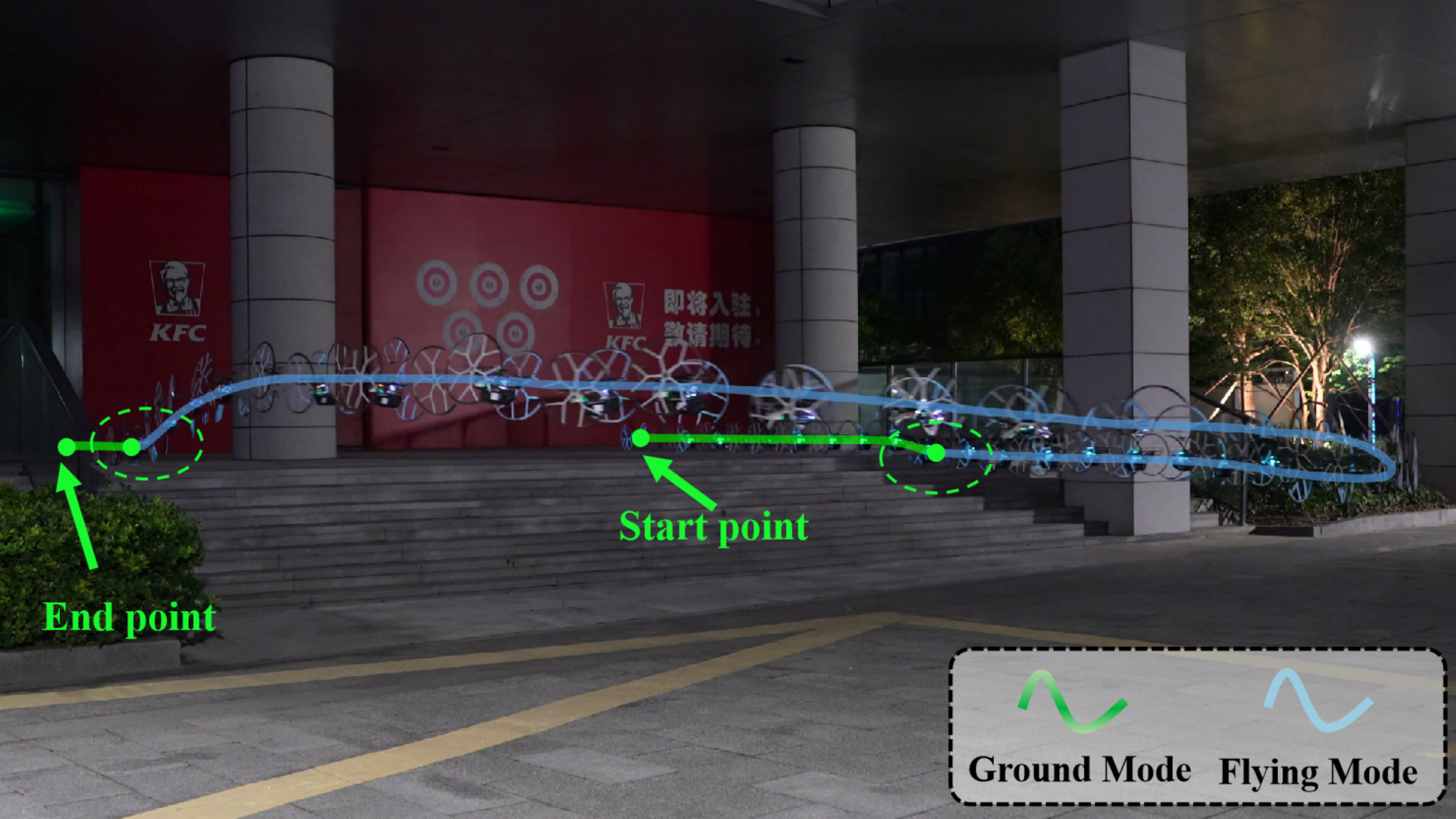}
    \caption{Real-world autonomous air-ground mode switching in a stair-like environment. Green and blue indicate ground and aerial locomotion, respectively.}
    \label{fig:real_switching_demo}
	\vspace{-0.8em}
\end{figure}

As shown in Fig.~\ref{fig:real_switching_demo}, the TABV successfully completes both Ground-to-Air and Air-to-Ground transitions while maintaining reference tracking across different locomotion modes. Using the same event definitions as in simulation, the measured switching delay for Ground Support Loss is $20.0\,\mathrm{ms}$, which is comparable to the simulated result of $28\pm12\,\mathrm{ms}$. For the Air-to-Ground transition, the switching delay under the Ground Recovery definition is $240.1\,\mathrm{ms}$, compared with $262\pm53\,\mathrm{ms}$ in simulation. 
\subsection{Integrated System Evaluation}
To evaluate the complete air-ground control framework, we conduct a long-horizon real-world experiment on the physical TABV, where the robot follows a reference trajectory and autonomously switches between ground and aerial locomotion using the proposed mode selector. As shown in Fig.~\ref{fig:HeadFigure}, the TABV completes a \(101\,\mathrm{m}\) air-ground trajectory with multiple autonomous mode transitions and achieves a position RMSE of \(0.08\,\mathrm{m}\), demonstrating reliable long-horizon trajectory tracking and mode coordination.

\section{Conclusion}\label{sec:Conclusion}

This paper presents a learning-based air-ground control framework for passive-wheeled terrestrial-aerial bimodal vehicles. The framework combines independently trained aerial and ground policies with a temporal mode selector, enabling autonomous air-ground switching under limited onboard perception. Simulation and real-world experiments demonstrate robust ground trajectory tracking across diverse terrains and reliable mode switching. The complete system achieves a position RMSE of \(0.08\,\mathrm{m}\) over a \(101\,\mathrm{m}\) real-world air-ground trajectory. Future work will focus on more complex and dynamic environments.
\bibliography{references}
\end{document}